# Multi Scale Temporal Graph Networks for Skeleton-Based Action Recognition

Tingwei Li[1], Ruiwen Zhang[2], Qing Li[1]

[1]Department of Automation Tsinghua University, Beijing, China
[2]Department of Computer Science Tsinghua University, Beijing, China

## ABSTRACT

*Graph convolutional networks (GCNs) can effectively capture the features of related nodes and improve the performance of model. More attention is paid to employing GCN in Skeleton-Based action recognition. But existing methods based on GCNs have two problems. First, the consistency of temporal and spatial features is ignored for extracting features node by node and frame by frame. To obtain spatiotemporal features simultaneously, we design a generic representation of skeleton sequences for action recognition and propose a novel model called Temporal Graph Networks (TGN). Secondly, the adjacency matrix of graph describing the relation of joints are mostly depended on the physical connection between joints. To appropriate describe the relations between joints in skeleton graph, we propose a multi-scale graph strategy, adopting a full-scale graph, part-scale graph and core-scale graph to capture the local features of each joint and the contour features of important joints. Experiments were carried out on two large datasets and results show that TGN with our graph strategy outperforms state-of-the-art methods.*

## 1. INTRODUCTION

Human action recognition is a meaningful and challenging task. It has widespread potential applications, including health care, human-computer interaction and autonomous driving. At present, skeleton data is more often used for action recognition because skeleton data is robust to the noise of background and different viewpoints compared to video data. Skeleton-based action recognition are mainly based on deep learning methods like Recurrent Neural Networks (RNNs), Convolutional Neural Networks (CNNs) and GCNs [3, 8, 10, 12, 13, 15, 17, 18,]. RNNs and CNNs generally process the skeleton data into vector sequence and image respectively. These representing methods cannot fully express the dependencies between correlated joints. With more researches on GCN, [12] first employs GCN in skeleton-based action recognition and inspires a lot of new researches [7, 10, 15, 17, 18].

The key of action recognition based on GCN is to obtain the temporal and spatial features of an action sequence through graph [7, 10, 18]. In the skeleton graph, skeleton joints transfer into node and the relations between joints are represented by edges. As shown in Fig. 1, in most previous work, there are more than one graphs, a node only contains spatial features. In this case, GCN extracts spatial features frame by frame, then Temporal Convolutional Network (TCN) extracts temporal features node by node. But, features of a joint in an action is not only related to other joints intra frames but also joints inter frames. As a result, existing methods split this consistency of spatiotemporal features. To solve this problem, we propose TGN to capture spatiotemporal features simultaneously, as shown in Fig. 2, each node composes a joint of all frames and contains both spatial and temporal feature in the graph, thus TGN obtains spatiotemporal features by processing all frames of each joint simultaneously.

Besides, the edges of skeleton graph mainly depend on the adjacency matrix A, which is related to the physical connections of joints [12,17,18]. GCN still have no effective adaptive graph mechanism to establish a global connection through the physical relations of nodes, such as a relation between head and toes. GCN can only obtain local features, such as a relation between head and neck. In this paper, a multi-scale graph strategy is proposed, which adopts different scale graphs in different network branches, as a result, physically unconnected information is added to help network capture the local features of each joint and the contour features of important joints.

The major contributions of this work lie in three aspects:

(1) This paper proposes a feature extractor called Temporal Graph Network (TGN) to obtain spatiotemporal features simultaneously, and this extractor can be fitted in and perform better than most skeleton-based action recognition models based on GCN.
(2) We devisea multi-scale graph strategy for optimization of graph to capture both the local features and the contour features.
(3) Combining the multi-scale graph strategy with TGN, we propose Multi-scale Temporal Graph Network (MS-TGN) which outperforms state-of-the-art methods on two large scale datasets for skeleton-based action recognition.

## 2. RELATED WORKS

### 2.1. Action Recognition based on Skeleton

The methods of skeleton-based action recognition can be classified into two kinds, using handcrafted features to model human bodies and using deep learning methods respectively. Using handcrafted features [5, 14] are quite complex and more suitable for small and medium-sized datasets. Deep learning methods are based on three models, RNNs, CNNs and GCNs. RNN-based methods [3, 19, 22] model temporal dependencies over sequences from skeletons. CNN-based approaches [2, 8] usually transfer the skeleton data as an image. For these methods cannot express a meaningful operator in the vertex domain. Recently, researchers tend to GCN [10, 12, 15, 18] and build operators in the non-Euclidean space.

### 2.2. Graph Convolutional Networks

In action recognition task, the principle of constructing GCN on the graph generally follows the spatial perspective [1], where the convolutional filters are applied directly to the graph nodes. Several classic methods in this task are proposed in recent two years. [12] embeds skeleton sequences into several graphs where joints in a frame of a sequence make up nodes of a graph and relations between joints are spatial edges of a graph. [18] gives a two-stream GCN architecture, which takes the second-order information (bones) into consideration and employs graph adaptiveness to optimize the adjacent matrix. [17] proposes graph regression based on GCN to exploit the dependencies of each joints. [15] employs Neural Architecture Search (NAS) to existing model to automatically design GCN.

## 3. OUR METHODS

### 3.1. Temporal Graph Networks

The skeleton data of an action can be described as $X = \{x_{c,v,t}\}_{C \times V \times T}$, where $T$ is the number offrames, $V$ is number of joints in a frame, $C$ is number of channels in a joint and $x_{c,v,t}$ represents the skeleton data of joint $v$ in the frame t with $c$ channels .$X_i$ is the sequence $i$. Previous methods construct $T$ graphs and each graph has $V$ nodes, as shown in Fig. 1, where the node set $N = \{x_{v,t}|, v = 1,2, ... V, t = 1,2 ... T\}_C$ has joint $v$ in $tth$ frame. It means a frame is represented as one graph and there are totally $T$ graphs. The size feature of a node is $C$. GCN is used to obtain spatial features from each graph, then outputs of GCN are fed into TCN to extract temporal features.

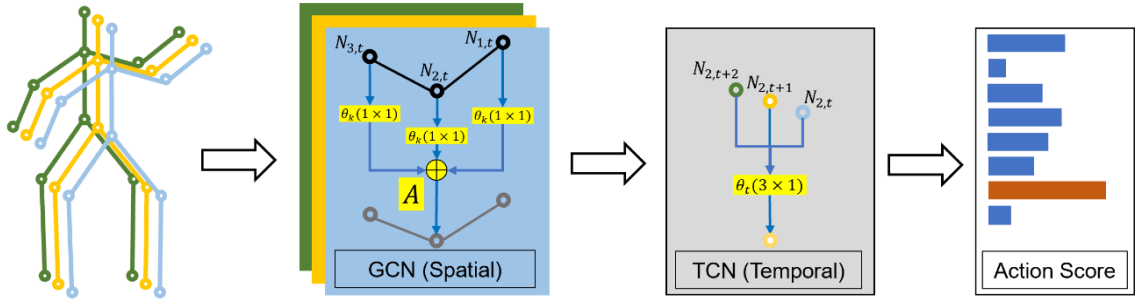

Figure 1. A node in any graph represents data of a joint in a certain frame. GCN extracts spatial features frame by frame, and TCN extracts temporal features node by node.

We redefine graph and propose TGN. Compared to $T$ graphs, we only have one graph with $V$ nodes, as seen in Fig. 2. In the graph, the node set $N = \{x_v|v = 1,2 ... T\}_{C \times V}$ has the joint $v$ in all frame. The size of feature of a node is $C \times V$. Compared with series methods using GCN and TCN alternately, we only use one GCN block to realize the extraction of spatiotemporal features.

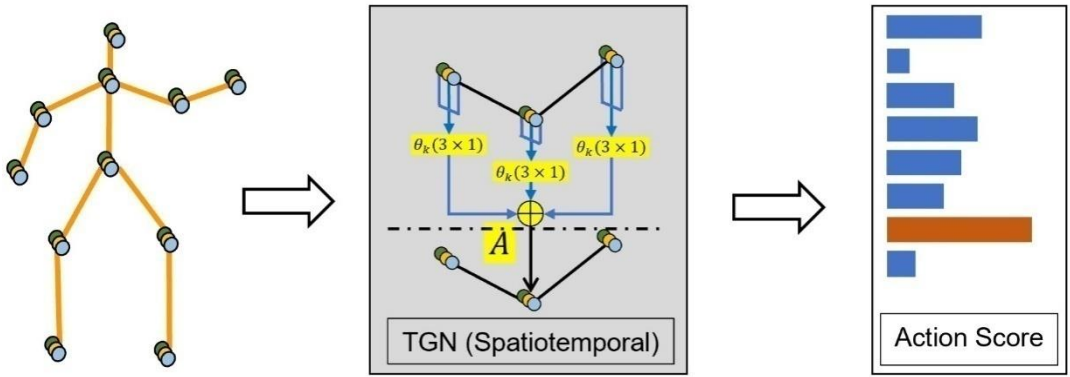

Figure 2. Each node represents data of a joint in all frames so temporal information is also contained. Compared with Fig1, TGN extracts temporal and spatial features simultaneously.

In a basic TGN block, each node already has temporal information and spatial information, therefore, in the process of graph convolution, temporal and spatial features can be calculated simultaneously. The output value for a single channel at the spatial location $N_v$ can be written as Eq. 2.

$$S_v = \{n_j|n_j \in S(N_v, h)\} \qquad (1)$$

$$F_o(N_v) = \sum_{j=0}^{k}(F_i(S_v(j)) \times w(j)) \quad (2)$$

Where $N_v$ is node $v$. $s(N_v, h)$ is a sampling function used to find node set $n_j$ adjacent to the node $N_v$, $F_i$ maps nodes to feature vector, $w(h)$ is weights of CNN whose kernel size is $1 \times t$, $F_o(N_v)$ is output of $N_v$. Eq.2 is a general formula among most GCN-based models of action recognition, as it was used to extract spatial features in a graph. our method can be adapted to existing methods by changing graph structure of this methods.

## 3.2. Multi-Scale Graph Strategy

Dilated convolution can obtain ignored features such as features between unconnected points in an image by over step convolution. Inspired of it, we select different expressive joints to form different scale graphs for convolution. Temporal features of joints with larger motion space are more expressive. Joints in a body generally have different relative motion space. For example, the elbow and knee can move in larger space compared to the surrounding joints like shoulder and span. In the small-scale graph, there are less but more expressive nodes so there is a larger receptive field. In large-scale graph, there are more but less expressive nodes so the receptive field is smaller.

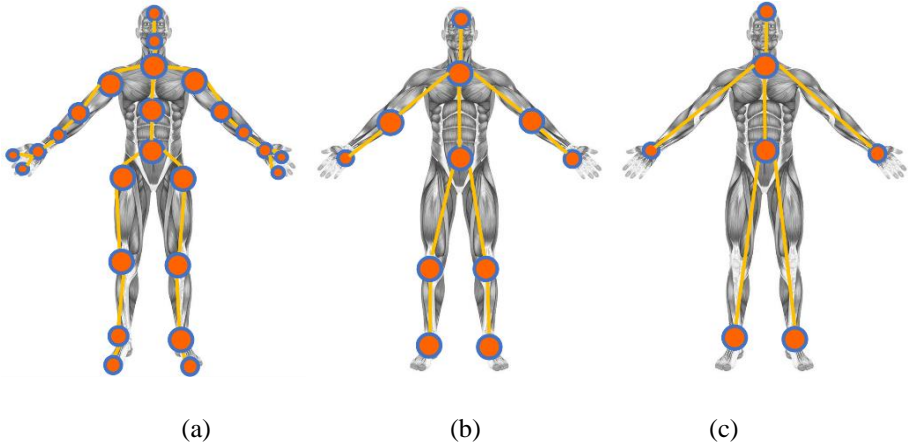

(a)    (b)    (c)
Figure 3.  (a) full-scale graph, (b) part-scale graph, (c) core-scale graph

We design three different scale graphs based on NTU-RGB+D datasets, as shown in Fig. 3. Full-scale graph in Fig. 3(a) has all 25 nodes and can obtain local features of each joint for its small receptive field. Part-scale graph in Fig. 3(b) is represented by only 11 nodes. In this case, receptive field becomes larger so it tends to capture contour information. Fig. 3(c) is core-scale graph with only seven nodes. It has largest convolution receptive field, although it ignores the internal state of the limbs, it can connect the left and right limbs directly, so the global information can be obtained. Through different scale graphs, GCN can capture local features, contour features and global features respectively.

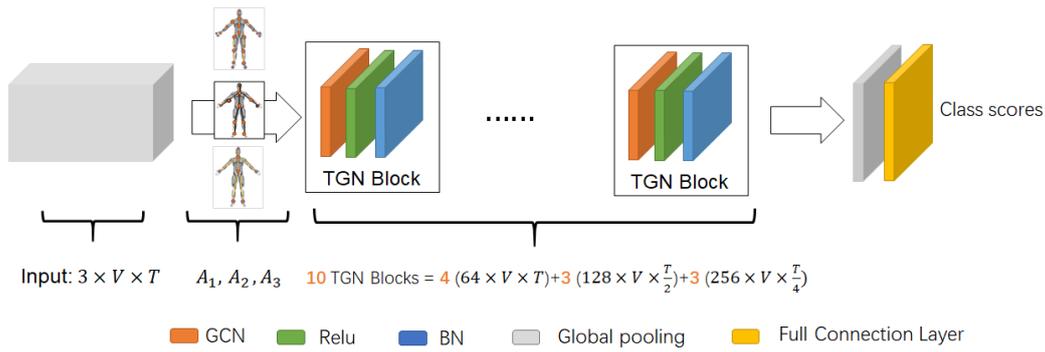

Figure 4. The network architecture of MS-TGN. Multi-scale graphs are displayed by different adjacency matrices $A_1, A_2$ and $A_3$.

Based on the former, we combine TGN with a multi-scale graph strategy to get the final model MS-TGN, as shown in Fig. 4. The model consists of 10 TGN layers, each layer uses $3 \times 1$ convolutional kernel to extract the temporal features of each node, and a fully-connected layer to classify based on the extracted feature.

## 4. EXPERIMENT

### 4.1. Datasets and Implementation Details

NTU RGB+D. This dataset is large and widely used. It contains 3D skeleton data collected by Microsoft's kinetics V2 [16] and has 60 classes of actions and 56,000 action sequences, with 40 subjects are photographed by three cameras fixed at 0◦, 45◦ and 45◦, respectively. Each sequence has several frames and each frame is composed of 25 joints. We adopt the same method [16] to carry out the cross-view (cv) and cross-subject (cs) experiments. In the cross-view experiments, the training data is 37,920 action sequences with the view at 45◦ and 0◦, and the test data is 18,960 action sequences with the view at 45◦ in the cross-subject experiments, the training data is action sequences performed by 28 subjects, and the test data contains 16,560 action sequences performed by others. We use the top-1 accuracy for evaluation.

Kinetics Skeleton. Kinetics [6] is a video action recognition dataset obtained from the video on YouTube. Kinetics Skeleton employs OpenPose [23] estimation toolbox to detect 18 joints of a skeleton. It contains 400 kinds of actions and 260,232 sequences. The training data consists of 240,436 action sequences, and the test data is the remaining 19,796 sequences. We use the top-1 and top-5 accuracies for evaluation.

Unless otherwise stated, all models proposed employ strategies following. The number of channels is 64 in the first four layers, 128 in the middle three layers and 256 in the last three layers. SGD optimizer with Nesterov accelerated gradient is used for gradient descent and different learning rate adjustment strategies are designed for different data. The mini-batch size is 32 and the momentum is set to 0.9. All skeleton sequences are padded to T = 300 frames by replaying the actions. Inputs are processed with normalization and translation as [18].

## 4.2. Ablation Experiments

### 4.2.1. Feature Extractor: TGN

Table 1. Effectiveness of our TGN module on NTU RGB+D dataset in terms of accuracy.

| Model | TGN | X-sub(%) | X-view(%) |
|---|---|---|---|
| ST-GCN[12] |  | 81.6 | 88.8 |
| ST-GCN[12] | √ | **82.3** | **90.8** |
| 2s-AGCN[18] |  | 88.5 | 95.1 |
| 2s-AGCN[18] | √ | **89.0** | **95.4** |
| Js-Ours |  | 86.0 | 93.7 |
| Js-Ours | √ | **86.6** | **94.1** |
| Bs-Ours |  | 86.9 | 93.2 |
| Bs-Ours | √ | **87.5** | **93.9** |

ST-GCN [12] and 2s-AGCN [18] are representative models utilizing GCN and TCN alternatively and were chosen as baselines. We replace GCN&TCN with TGN in these two models and keep adjacency matrix construction strategies unchanged. The experimental results on the two datasets are listed in Table 1 and Table 2. From Table 1, the original performance of ST-GCN increases to 82.3% and 90.8% on X-sub and X-view, and the accuracy of 2s-AGCN increases by 0.55% on average. From Table 2, accuracies of two models are both improved. In conclusion, TGN is sufficiently flexible to be used as a feature extractor and performs better than methods based on GCN & TCN.

Table 2. Effectiveness of our TGN module on Kinetics dataset in terms of accuracy.

| Model | TGN | Top-1(%) | Top-5(%) |
|---|---|---|---|
| ST-GCN[12] |  | 30.7 | 52.8 |
| ST-GCN[12] | √ | **31.5** | **54.0** |
| 2s-AGCN[18] |  | 36.1 | 58.7 |
| 2s-AGCN[18] | √ | **36.7** | **59.5** |
| Js-Ours |  | 35.0 | 93.7 |
| Js-Ours | √ | **35.2** | **94.1** |
| Bs-Ours |  | 33.0 | 55.7 |
| Bs-Ours | √ | **33.3** | **56.2** |

### 4.2.2. Multi-Scale Graph Strategy

Based on section 4.2.1, we construct the full-scale graph containing all joints of the original data, and the part-scale graph containing 11 joints in NTU+RGB-D dataset. In Kinetics Skeleton dataset, the full-scale graph contains all nodes and the part-scale graph contains 11 joints. We evaluate each scale graph, and Table 3 and Table 4 show the experimental results on the two datasets. In detail, adding a part-scale graph increases accuracy by 1.5% and 0.45%, respectively, adding a core-scale graph increases accuracy by 0.3% and 0.2%, respectively. A core-scale graph provides the global features of the whole body, a part-scale graph provides the contour features of the body part, and a full-scale graph provides the local features of each joint. By feature fusion, the model obtains richer information and performs better.

Table 3. Effectiveness of Multi-scale Graph on NTU RGB+D dataset in terms of accuracy.

| Full-Scale | Part-Scale | Core-Scale | X-sub(%) | X-view(%) |
|---|---|---|---|---|
| √ | | | 89.0 | 95.2 |
| | √ | | 86.0 | 94.0 |
| | | √ | 85.6 | 93.3 |
| √ | √ | | 89.2 | 95.7 |
| √ | √ | √ | 89.5 | 95.9 |

Table 4. Effectiveness of Multi-scale Graph on Kinetics dataset in terms of accuracy.

| Full-Scale | Part-Scale | Core-Scale | Top-1(%) | Top-5(%) |
|---|---|---|---|---|
| √ | | | 36.6 | 59.5 |
| | √ | | 35.0 | 55.9 |
| | | √ | 33.8 | 54.6 |
| √ | √ | | 36.9 | 59.9 |
| √ | √ | √ | 37.3 | 60.2 |

### 4.3. Comparison with State-of-the-Art Methods

We compare MS-TGN with the state-of-the-art skeleton-based action recognition methods on both the NTU RGB+D dataset and the Kinetics-Skeleton dataset. The results of NTU RGB+D are shown in Table 5. Our model performs the best in cross-view experiment on NTU RGB+D, and has the highest top-1 accuracy on Kinetics Skeleton dataset as listed in Table 6.

Table 5. Performance comparisons on NTU RGB+D dataset with the CS and CV settings.

| Method | Year | X-sub(%) | X-view(%) |
|---|---|---|---|
| HBRNN-L[7] | 2015 | 59.1 | 64.0 |
| PA LSTM[16] | 2016 | 62.9 | 70.3 |
| STA-LSTM[21] | 2017 | 73.4 | 81.2 |
| GCA-LSTM[22] | 2017 | 74.4 | 82.8 |
| ST-GCN[12] | 2018 | 81.5 | 88.3 |
| DPRL+GCNN[25] | 2018 | 83.5 | 89.8 |
| SR-TSL[19] | 2018 | 84.8 | 92.4 |
| AS-GCN[10] | 2019 | 86.8 | 94.2 |
| 2s-AGCN[18] | 2019 | 88.5 | 95.1 |
| VA-CNN[26] | 2019 | 88.7 | 94.3 |
| SGN[27] | 2020 | 89.0 | 94.5 |
| **MS-TGN(ours)** | - | **89.5** | **95.9** |

Table 6. Performance comparisons on Kinetics dataset with SOTA methods.

| Method | Year | Top-1(%) | Top-5(%) |
|---|---|---|---|
| PA LSTM[16] | 2016 | 16.4 | 35.3 |
| TCN[12] | 2017 | 20.3 | 40.0 |
| ST-GCN[12] | 2018 | 30.7 | 52.8 |
| AS-GCN[10] | 2019 | 34.8 | 56.6 |
| 2s-AGCN[18] | 2019 | 36.1 | 58.7 |
| NAS[15] | 2020 | 37.1 | 60.1 |
| **MS-TGN(ours)** | - | **37.3** | **60.2** |

Besides, our model reduces the computational work and parameters, as listed in Table 7. It means our model is simpler and has a better ability for modelling spatial and temporal features. Why does our model have fewer parameters and calculation but perform better? In our designed graph, each node denotes all frame data of a joint, which brings two advantages: (1) TGN extractor only contains GCN without TCN, which reduces the parameters and calculation. (2) Instead of extracting alternately, TGN can extract spatial and temporal features at the same time to strengthen consistence of the spatial and temporal features.

Table 7.  Comparisons of the cost of computing with state-of-the-arts. The #Params and FLOPs are calculated by the tools called THOP (PyTorch-OpCounter) [24].

| Method | Year | X-sub(%) | #Params(M) | #FLOPs(G) |
|---|---|---|---|---|
| ST-GCN[12] | 2018 | 81.6 | 3.1 | 15.2 |
| AS-GCN[10] | 2019 | 86.8 | 4.3 | 17.1 |
| 2s-AGCN[18] | 2019 | 88.5 | 3.5 | 17.4 |
| NAS[15] | 2020 | 89.4 | 6.6 | 36.6 |
| **MS-TGN(ours)** | - | **89.5** | **3.0** | **15.0** |

## 5. CONCLUSIONS

In this paper, we propose MS-TGN model which mainly has two innovations: a model called TGN to extract temporal and spatial features simultaneously and a multi-scale graph strategy to obtain both the local features and the contour features. On two large-scale datasets, the proposed MS-TGN achieves the state-of-the-art accuracy with the least parameters and calculation.

## ACKNOWLEDGEMENT

This work is sponsored by the National Natural Science Foundation of China No. 61771281, the "New generation artificial intelligence" major project of China No. 2018AAA0101605, the 2018 Industrial Internet innovation and development project, and Tsinghua University initiative Scientific Research Program.

## AUTHORS

**Li Tingwei** is a master's student in Department of Automation in Tsinghua University, Beijing, China. She researches action recognition and architecture.

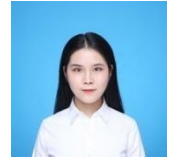

**Zhangruiwen** is currently pursuing the masterdegree with the Computer Science in Tsinghua University. He is studying auto-driving and action recognition.

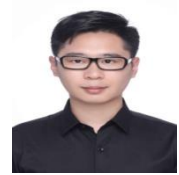

**Qing Li, PhD**, is a professor of the Department of Automation, Tsinghua University, P.R. China. He has taught at Tsinghua University since 2000 with major research interests in system architecture, enterprise modelling and system performance evaluation.

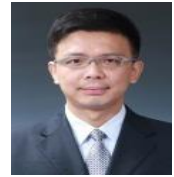